\documentclass[10pt,twocolumn,letterpaper]{article}

\usepackage{wacv}              

\usepackage{graphicx}
\usepackage{amsmath}
\usepackage{amssymb}
\usepackage{booktabs}
\usepackage{xcolor, colortbl}
\usepackage{multirow}
\usepackage{multicol}
\usepackage{cuted}
\usepackage{capt-of}
\usepackage{subcaption}
\usepackage{notoccite}
\usepackage{placeins}

\usepackage[pagebackref,breaklinks,colorlinks]{hyperref}
\usepackage{hypcap}

\usepackage[capitalize]{cleveref}
\crefname{section}{Sec.}{Secs.}
\Crefname{section}{Section}{Sections}
\Crefname{table}{Table}{Tables}
\crefname{table}{Tab.}{Tabs.}

\def\wacvPaperID{*****} 
\def\confName{WACV}
\def\confYear{2025}

\begin{document}

\title{Benchmarking Image Similarity Metrics for Novel View Synthesis Applications}

\author{
  \begin{tabular}{ccc}
  Charith Wickrema & Sara Leary & Shivangi Sarkar\\
  \end{tabular} \\
  \begin{tabular}{cccc}
  Mark Giglio & Eric Bianchi & Eliza Mace & Michael Twardowski
  \end{tabular} \\
  The MITRE Corporation \\
  {\tt\small \{cwickrema, sleary, ssarkar, mgiglio, ebianchi, emace, mtwardowski\}@mitre.org}
}

\maketitle

\vspace{-1.5cm} 

\begin{strip}
  \centering
  \includegraphics[width=0.8\textwidth]{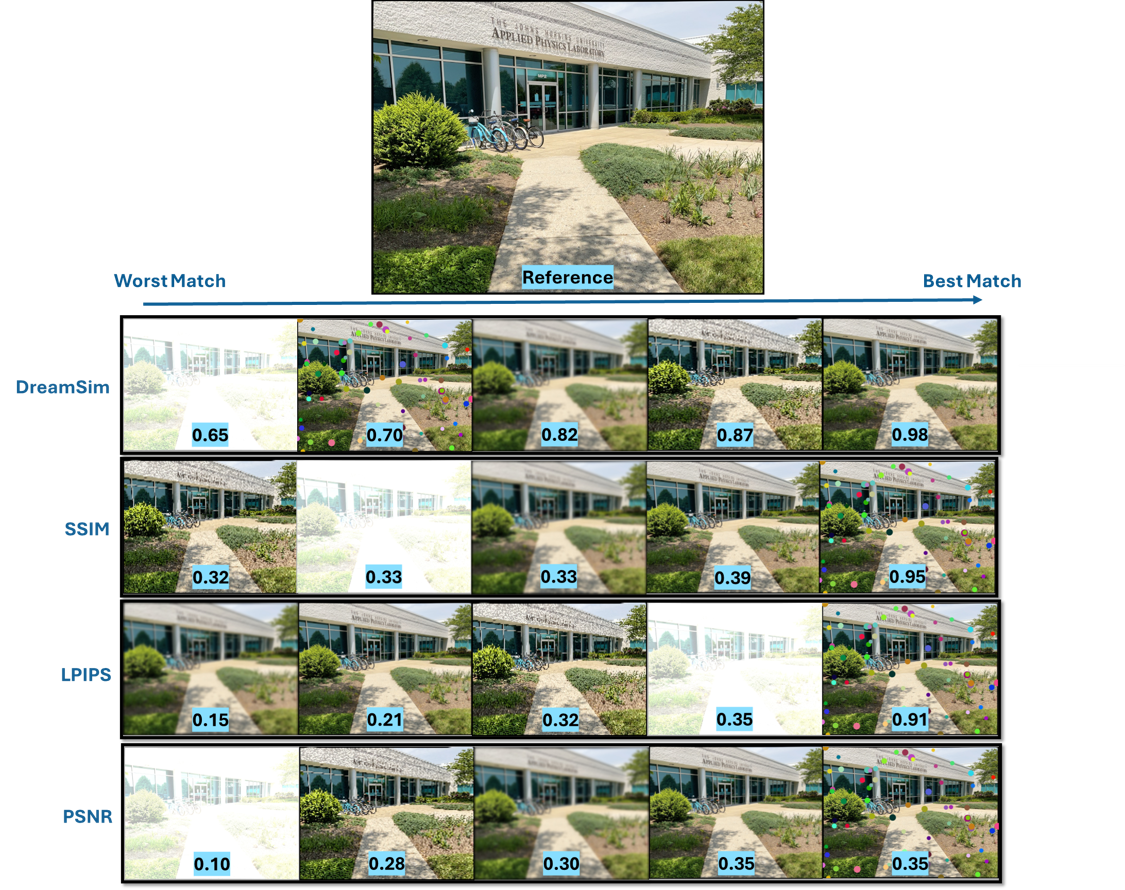}
  \captionof{figure}{\textbf{Which metric most effectively scores the similarity of these images to the reference?} Our paper assesses commonly used image similarity metrics on how well they evaluate images with corruptions commonly found in NVS renders. Each row represents a sequence of images scored by a different metric, where the scores are normalized from 0 (lowest similarity) to 1 (highest similarity).}
  \label{fig:fig1_metrics_comparison}
\end{strip}

\begin{abstract}
  Traditional image similarity metrics are ineffective at evaluating the similarity between a real image of a scene and an artificially generated version of that viewpoint \cite{eval_qual_mets_nerf, perceptualqualassessment, subjective_objective_eval,nerf_nqa}. 
  Our research evaluates the effectiveness of a new, perceptual-based similarity metric, DreamSim \cite{dreamsim}, and three popular image similarity metrics: Structural Similarity (SSIM), Peak Signal-to-Noise Ratio (PSNR), and Learned Perceptual Image Patch Similarity (LPIPS) \cite{ssim, lpips} in novel view synthesis (NVS) applications.
  We create a corpus of artificially corrupted images to quantify the sensitivity and discriminative power of each of the image similarity metrics.
  These tests reveal that traditional metrics are unable to effectively differentiate between images with minor pixel-level changes and those with substantial corruption, whereas DreamSim is more robust to minor defects and can effectively evaluate the high-level similarity of the image. 
  Additionally, our results demonstrate that DreamSim provides a more effective and useful evaluation of render quality, especially for evaluating NVS renders in real-world use cases where slight rendering corruptions are common, but do not affect image utility for human tasks.
\end{abstract}

\footnotetext{Approved for Public Release; Distribution Unlimited. Public Release Case Number 24-2132}

\section{Introduction}
\label{sec:intro}
Novel view synthesis (NVS) algorithms aim to generate unique perspectives of a scene given a limited set of observed input views and a desired vantage point. 
State of the art NVS methods, such as Neural Radiance Fields (NeRFs) \cite{nerf} and 3D Gaussian Splatting (3DGS) \cite{3DGS}, have demonstrated remarkable progress in capturing complex scene geometry and generating novel, photorealistic views. 
As a result of these improved capabilities, NVS models have gained popularity across diverse computer vision application areas including: site models for humanitarian and disaster relief efforts, where timely image capture capabilities may be limited; law enforcement or first responder preparedness; and augmented reality/virtual reality applications \cite{program_baa,nerf_wild}. 
Effective application of NVS technology to such use cases necessitates that the algorithm outputs capture object- and scene-level content accurately. 

Despite these recent advancements, evaluating the performance of NVS models remains a challenge due to the complexity of quantifying human task-relevant properties. 
Commonly used metrics, such as SSIM, PSNR, and LPIPS, are considered low-level \cite{ssim,lpips}, as they primarily focus on individual pixel values or properties of small image portions, referred to as ``patches.'' 
While these metrics provide some valuable insights in tightly controlled lab settings \textemdash where little noise or pose-estimation error is introduced \textemdash empirical evidence \cite{eval_qual_mets_nerf,perceptualqualassessment,subjective_objective_eval,nerf_nqa} suggests that they do not suffice for assessing image utility for human-centric applications in real-world environments, with less consistent input data and outputs with varying levels of corruption. 
An ideal metric for NVS applications aligns with human judgement by scaling its score with image similarity and inversely to the level of corruption present in an image. 

In our research, we assess the effectiveness of popular low- and high-level image similarity metrics for evaluation of NVS renders for real-world applications.
\section{Related Work}

\label{sec:related_work}

\subsection{Novel View Synthesis (NVS)}
Most NVS models comprise a two-step process for generating images. 
First, a pose estimation step uses an algorithm, commonly a structure-from-motion based approach \cite{sfm_colmap}, to estimate camera positioning from input images. 
Second, a training step leverages the same images and their previously estimated camera positions as input and optimizes a model to minimize the per-pixel difference between generated images and ground truth reference images. 
Once trained, models can generate new images from novel viewpoints not included in the input data. 
Training an accurate model is highly dependent on the pose estimation step, which is more challenging in real world applications where data is often sparse or inconsistent \cite{nerf_wild,benchmarking_nerf_robustness}. 

\begin{figure}
  \centering
  \includegraphics[width=0.82\linewidth]{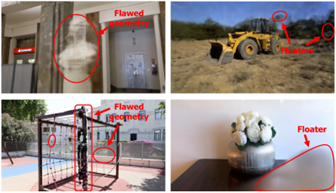}
  \caption{Examples of real NVS artifacts \cite{subjective_objective_eval}. Notably, NVS artifacts can affect the entire image as well as local patches of an image.}
  \label{fig:NVS_artifact_examples}
\end{figure}
Recently, the state of the art in this field has progressed significantly due to new training techniques used by NeRFs and 3DGS models. 
These new techniques, although they demonstrate a significant improvement, still produce errant outputs, such as artifacts within the rendered image due to misestimated scene geometry, such as those depicted in Figure \ref{fig:NVS_artifact_examples} \cite{subjective_objective_eval}. 
These artifacts can have a negative impact on the renders' perceived quality. 
Currently, NVS render quality is primarily judged using three image similarity metrics \cite{subjective_objective_eval}, which compare rendered views to ground truth images to produce a similarity score. 

\subsection{Image Similarity Metrics}
Metrics currently used for NVS, such as SSIM, PSNR, and LPIPS \cite{ssim, lpips}, were developed for use in the field of image compression, where they were used to measure the low-level similarity of reconstructed images as compared to their uncompressed source inputs to quantify the fidelity of the compression process. 
For this application, metrics that can differentiate very subtle changes in pixel or patch values are ideal. 
However, in NVS applications, slight differences between a rendered image and the reference image, due to changes in the environment, view angle, camera parameters, and artifacts, often result in scores inconsistent with human perception of quality. 
For example, quantifying the fidelity of the overall content, such as objects and their relative placement, has more utility for human use cases than measuring per-pixel differences. 

Recently, a higher-level perceptual similarity metric, DreamSim, has demonstrated strong alignment with human judgement on image similarity tasks \cite{dreamsim}. 
Due to its focus on higher-level features in an image, DreamSim has potential as a metric for the NVS field. 
It is worth noting that a preliminary assessment of DreamSim's performance indicated that the algorithm exhibits foreground clarity bias, meaning that artifacts impacting prominent objects in a scene are penalized more heavily than those exclusively affecting background scene content \cite{dreamsim}.

\section{Methods}
\label{sec:methods}

To quantify the efficacy of different metrics for NVS applications, we perform two experiments using artificially generated corruptions that emulate common NVS render artifacts, examples of which are shown in Figure \ref{fig:NVS_artifact_examples}. 

In the first experiment, we analyze the effects of common global corruptions applied to images from two datasets.  
We first use a subset of ImageNet-C \cite{imagenetC}, an open-source corrupted ImageNet dataset for benchmarking image classifier robustness. 
Second, we collect a series of images of scenes primarily composed of buildings. 
Each scene has several images captured from multiple angles for the purpose of NVS site modeling. 
These images are corrupted to emulate the artifacts often seen in NVS renders and for the purpose of this evaluation is referred to as NVS Scenes.
This dataset is pending approval for public release.  
\begin{table}
  \centering
  {\small{
  \begin{tabular}{|p{0.225\linewidth}|c|c|c|c|}
    \hline
    Dataset & Images & Corruptions & Severities & Total \\
    \hline
    NVS Scenes & 24 & 12 & 20 & 5,760 \\
    ImageNet-C & 10k & 15 & 5 & 750k \\
    \hline
  \end{tabular}
  }}
  \caption{A list of the number of images, corruptions, and severity levels applied to each dataset and the total number of combinations tested.}
  \label{tbl:datasets_overview}
\end{table}

In the second experiment, we test the effects of localized corruptions on downstream metrics utilizing the ImageNet-C dataset in two ways. 
First, we explore the effects of foreground versus background corruptions to explore DreamSim's foreground clarity bias \cite{dreamsim} on real imagery. 
Second, we test how metrics respond to image cropping by a small number of pixels; this corruption is almost imperceptible to the human eye and is not likely to impact the utility of NVS renders for human tasks. 

For both experiments, we normalize the results of each image similarity metric to be on a common scale of zero to one, where 1 is the highest score, indicating an image is identical to the reference, and 0 is the lowest possible score, indicating no similarity.  

\subsection{Global Corruptions}
\begin{figure}
  \centering
  \includegraphics[width=0.90\linewidth]{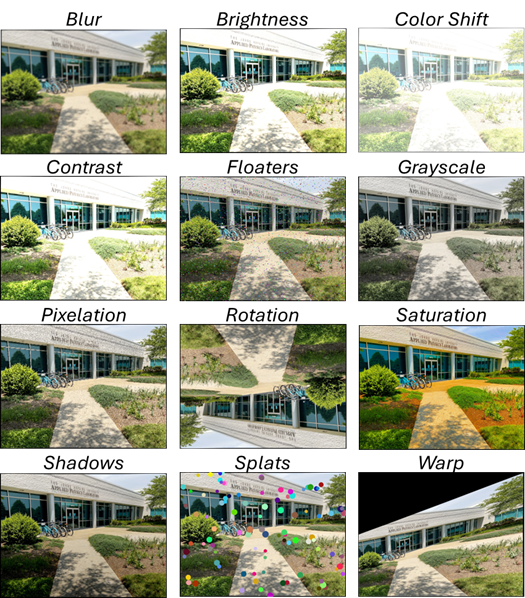}
  \caption{The 12 unique corruption types applied to the NVS Scenes dataset at level 10 intensity.}
  \label{fig:NVS_corruptions_example}
\end{figure}

\begin{figure}
  \centering
  \includegraphics[width=0.90\linewidth]{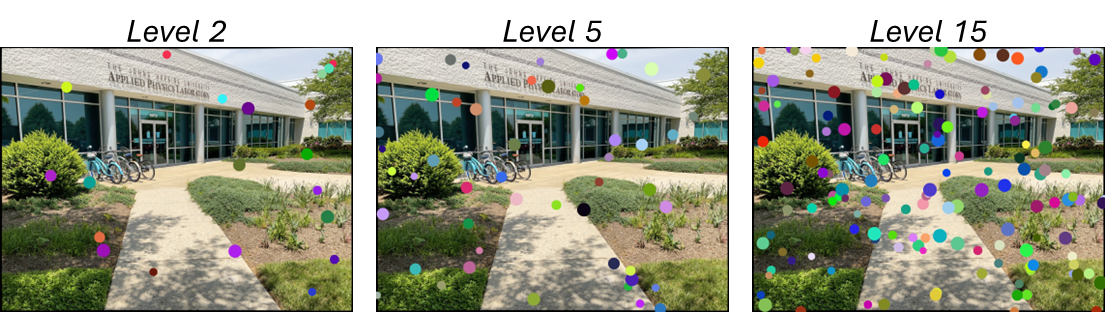}
  \caption{Example of the splat corruption at intensity levels 2, 5, and 15.}
  \label{fig:NVS_corruption_severity}
\end{figure}

We evaluate the effects of global image corruption with two different corruption datasets detailed in Table \ref{tbl:datasets_overview}. 
We implement a new set of 12 image corruption algorithms at 20 severity levels on our NVS Scenes dataset. 
We refer to the new corruption suite as ``NVS Corruptions,'' which simulates common corruptions typically seen in NVS model outputs, specifically: blur; brightness; color shift; contrast; floaters; grayscale; pixelation; rotation; saturation; shadows; splats; and warp. 
We apply these corruption algorithms to every image in NVS Scenes dataset at 20 severity levels. 
Examples of the NVS Corruptions are shown in Figure \ref{fig:NVS_corruptions_example} and an example of the severity levels is illustrated in Figure \ref{fig:NVS_corruption_severity} with splats corruption as an exemplar corruption.
This dataset is representative of real-world NVS tasks and lends itself to assessment of image similarity metrics over a limited variety of scenes.

\begin{figure}
  \centering
  \includegraphics[width=0.86\linewidth]{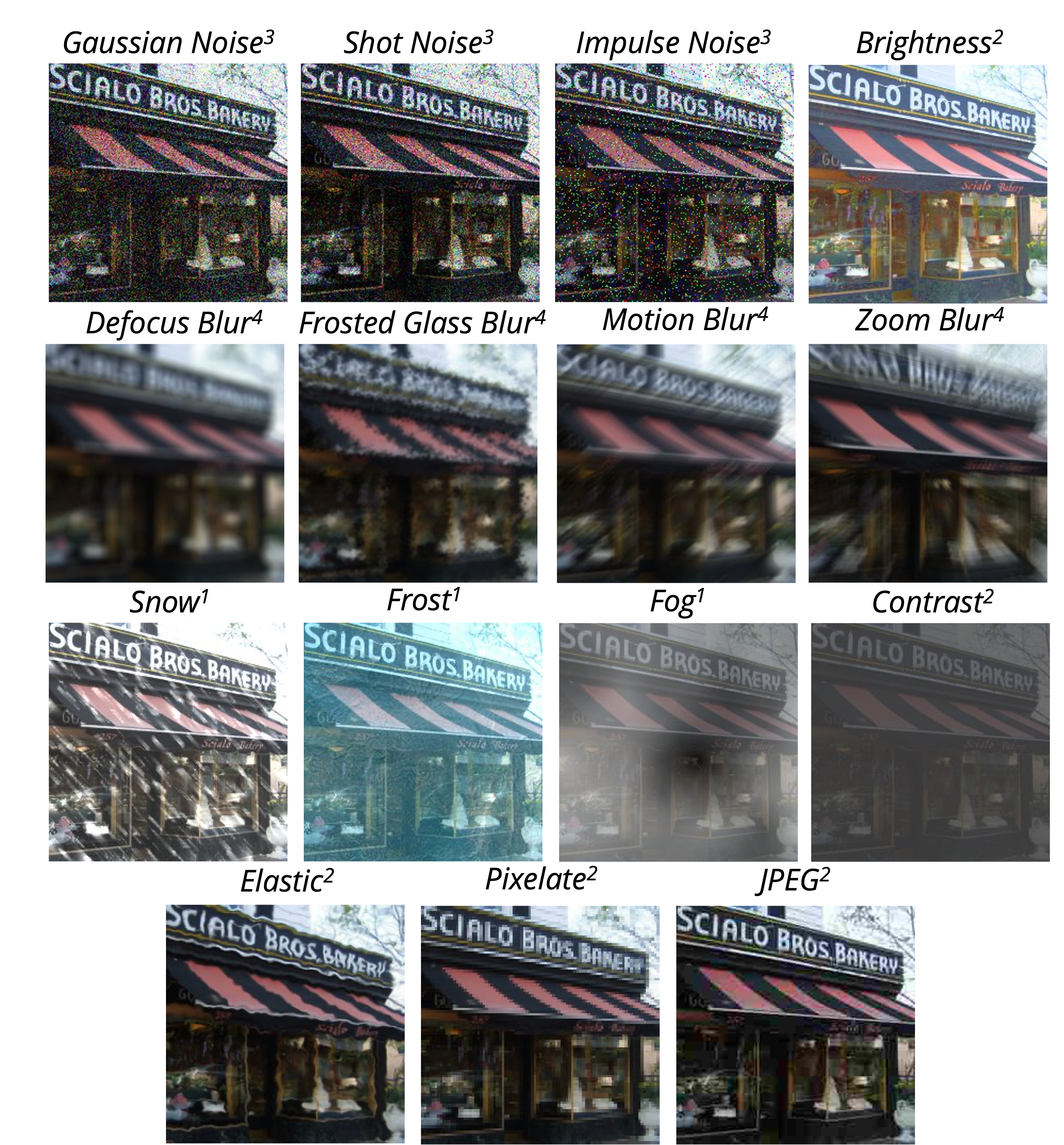}
  \caption{Examples of the corruptions included in the ImageNet-C dataset. Each corruption type's category is indicated with superscript numbers for these categories: weather (1), digital artifacts (2), noise (3), and blur (4).}
  \label{fig:ImageNetC_examples}
\end{figure}

\begin{figure}
  \centering
  \includegraphics[width=0.86\linewidth]{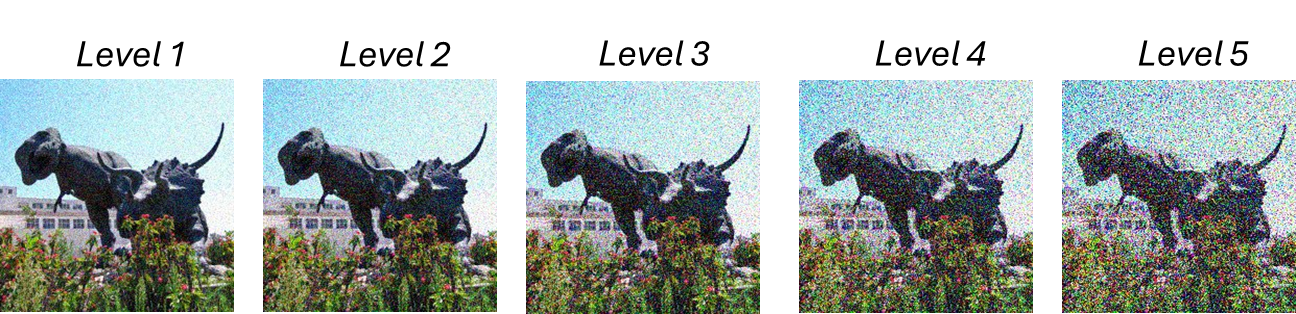}
  \caption{Illustration of each severity level of the Gaussian noise corruption applied to an image, from least severe (level 1) to most severe (level 5).}
  \label{fig:ImageNetC_severities}
\end{figure}

In addition to our NVS Scenes dataset, we evaluate image similarity metrics against a subset of ImageNet-C. 
Due to computational limitations, we curate 750k uniquely corrupted images from a random subset of 200 random classes from the original 1,000 classes of ImageNet \cite{imagenet}. 
ImageNet-C corruptions fall into four categories (weather, digital artifacts, noise, and blur);  an example of each type is shown in Figure \ref{fig:ImageNetC_examples}, and an example of each severity level is provided in Figure \ref{fig:ImageNetC_severities}. 
While the standard ImageNet-C dataset has less granularity in its severity levels and fewer NVS specific corruptions, its diversity and volume provide utility for our assessment. 

\subsection{Local Corruptions}
We apply our own two types of local corruptions to ImageNet-C uncorrupted images: foreground and background corruptions, which test both metric sensitivity to the corruptions themselves, as well as to the size of the applied corruption; and pixel cropping, in which we crop the images by a small number of pixels, simulating NVS renders where the camera view is off by a human imperceptible amount (5 to 10 pixels).  

\subsection{Foreground and Background Corruptions}
\begin{figure}
  \centering
  \vspace{-5pt}
  \includegraphics[width=0.65\linewidth]{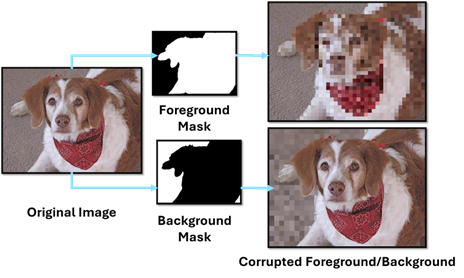}
  \caption{Foreground/Background Corruptions overview: original images are segmented and then corrupted at 20 levels of severity.}
  \label{fig:ForeBackground_pipeline}
\end{figure}

We leverage a segmentation pipeline to create foreground and background masks of images.
Our pipeline uses Lang-SAM \cite{langsam}, which is an integration of Grounding-DINO \cite{groundingdino} with Segment Anything \cite{segment_anything}. 
Lang-SAM intakes text-prompts \textemdash in our case, ``foreground'' and ``background'' \textemdash to create bounding boxes, which are then passed into a pretrained Segment Anything model that outputs segmentation masks of the target areas. 
As shown in Figure \ref{fig:ForeBackground_pipeline}, these masks are used to apply 10 of 12 of the corruptions from our NVS Corruption suite that are applicable to subsections of an image. 
Specifically, we exclude the rotation and warp corruptions because they alter the entire image. 
Finally, we manually validate the masks; the time intensive nature of this step limits the size of this evaluation dataset to approximately 500 images.

\subsection{Imperceptible Image Cropping}
\begin{figure}[h!]
  \centering
  \includegraphics[width=0.75\linewidth]{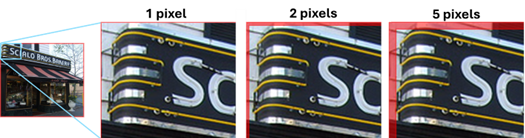}
  \caption{A zoomed-in illustration of a corner of an image shown where red indicates cropped pixels. ImageNet images were cropped around all sides by 1, 2, 5, and 10 pixels.}
  \label{fig:pixelcrop_example}
\end{figure}

In the pixel cropping corruption process, a small number of pixels are cropped out on each edge of approximately 2,000 ImageNet images. 
Each image is cropped evenly around all sides by 1, 2, 5, and 10 pixels and then run through each metric in comparison to its unaffected reference image. 
Pixel cropping examples are shown in Figure \ref{fig:pixelcrop_example}. 
The goal of testing this corruption is to quantify how each metric is affected by minimal corruptions that do not greatly impact overall scene understanding.
Additionally, pixel cropping corruption emulates the types of artifacts that NVS algorithms are prone to due to errors during their process of estimating camera positioning.

\section{Experimental Results and Discussion}
\label{sec:results}

\begin{figure*}[t!]
  \centering
  \includegraphics[width=0.8\textwidth]{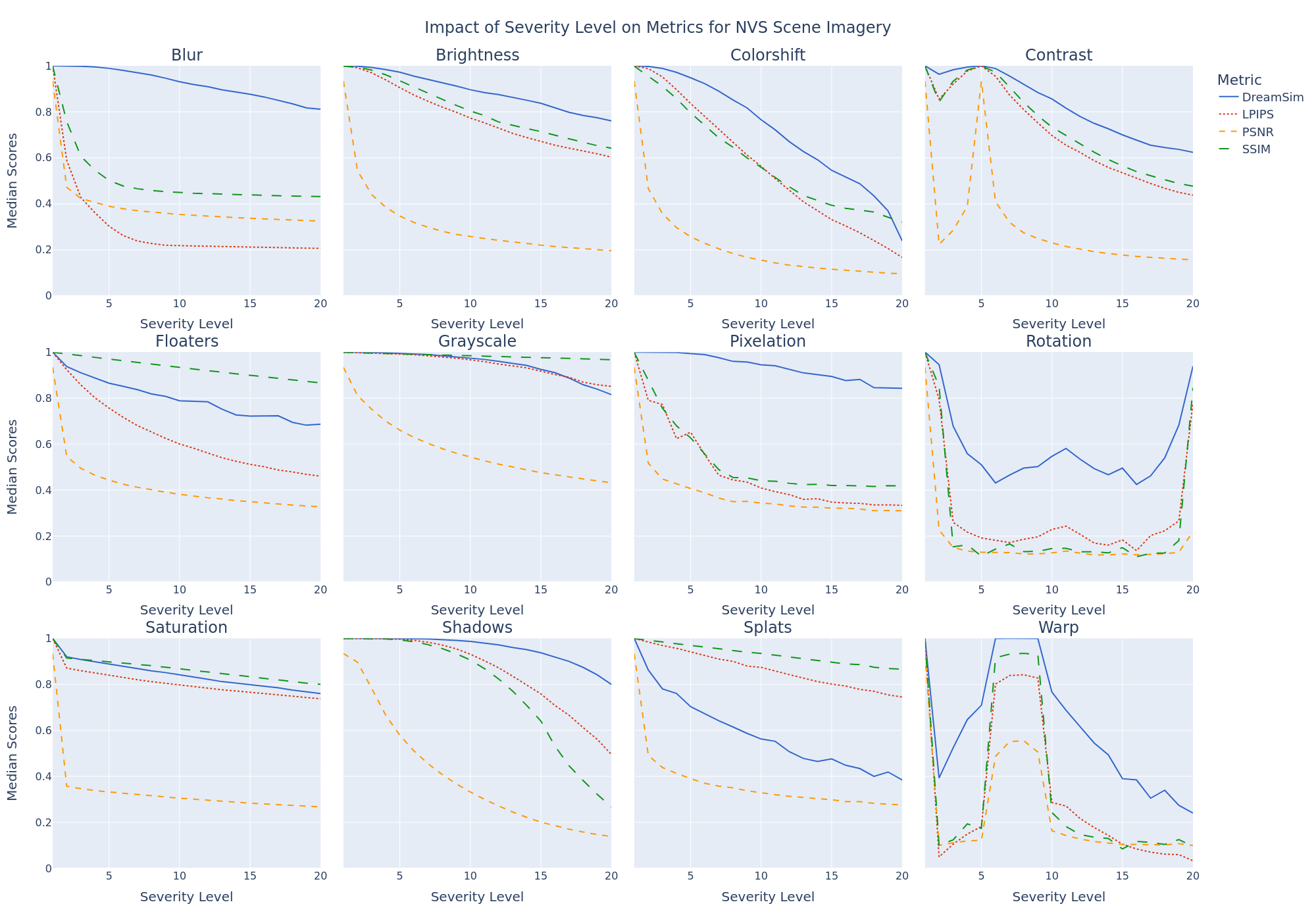}
  \caption{Median scores graphed across each severity level in our NVS rendered dataset. All metric scores are normalized to the same scale (0 to 1), where 1 indicates a perfectly identical image pair.}
  \label{fig:NVS_Corruption_Results}
\end{figure*}

\begin{figure}
  \centering
  \includegraphics[width=\linewidth]{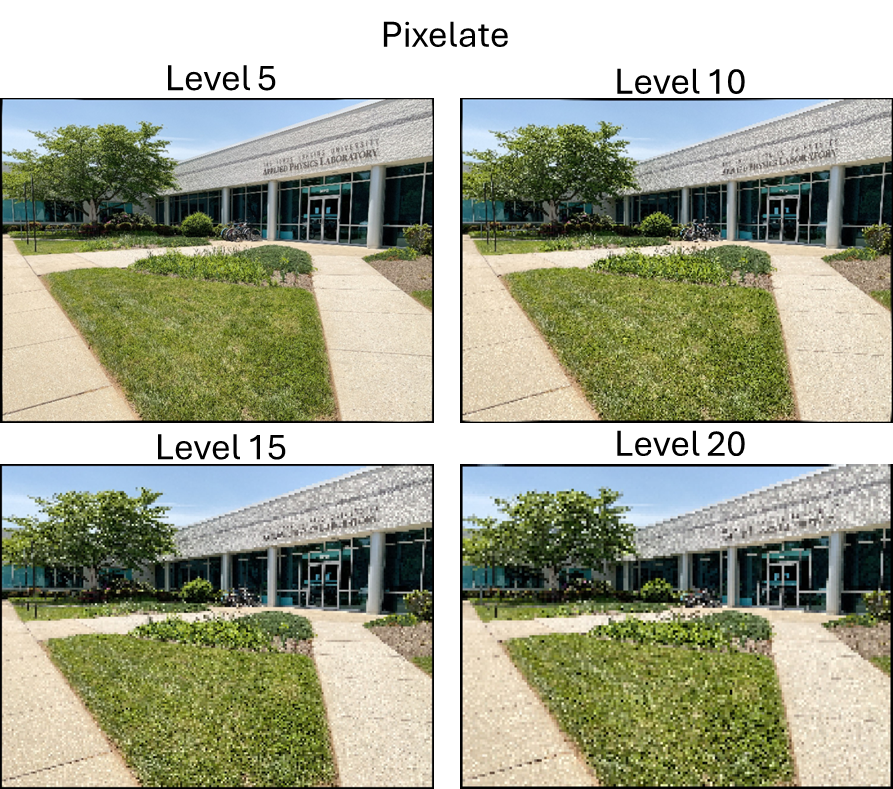}
  \caption{An example of the pixelation corruption at different severity levels.}
  \label{fig:pixelation_example}
\end{figure}

\begin{figure}
  \centering
  \includegraphics[width=\linewidth]{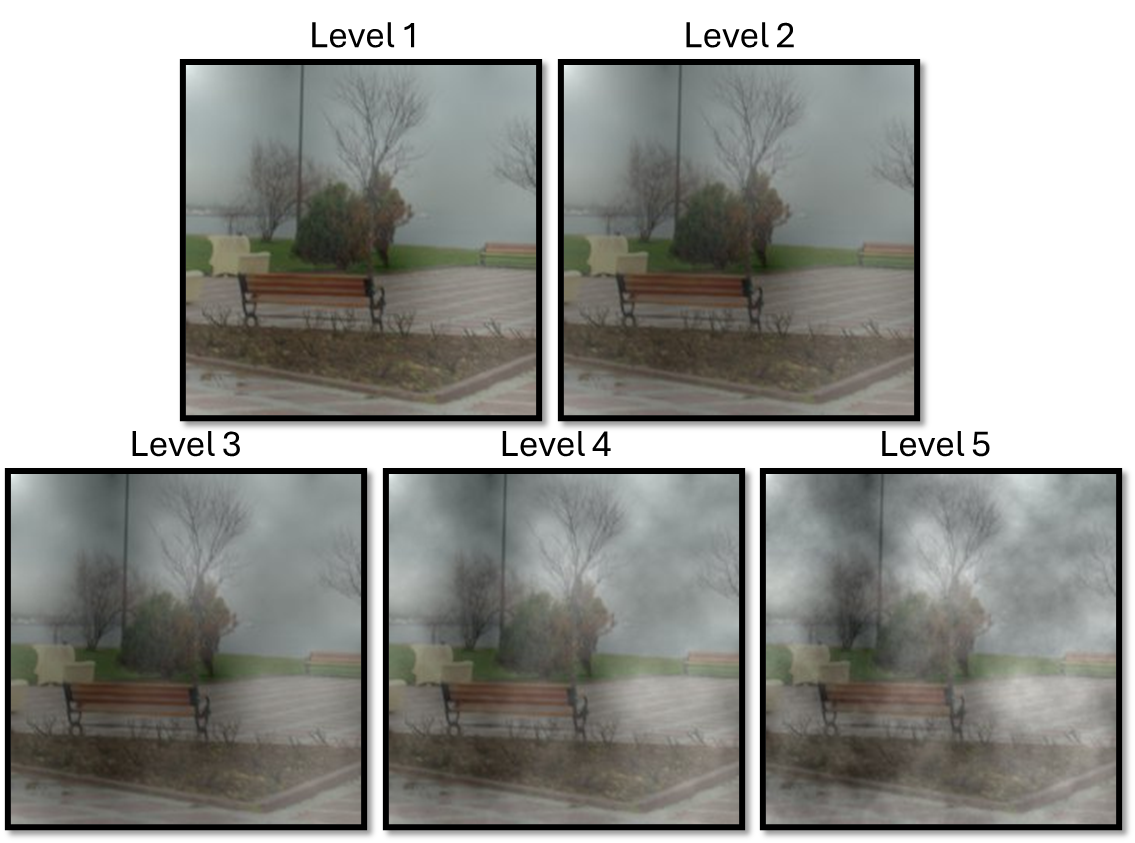}
  \caption{An example of the fog corruption, which had little perceptual difference between severity levels.}
  \label{fig:blur_example}
\end{figure}

\subsection{Global Corruptions}
The results of applying image similarity metrics to the NVS renders with global corruptions are shown in Figure \ref{fig:NVS_Corruption_Results}. 
These results depict that DreamSim gradually decreases its similarity score as the intensity of blur application increases. 
Conversely, SSIM, LPIPS, and PSNR all demonstrate a steep drop-off in score after minimal blurring is applied; this decreased score remains relatively consistent across all further blurring intensities, indicating that these metrics do not discriminate well between corruption levels. 
Not only does DreamSim consistently align with human judgement (examples shown in \ref{fig:fig1_metrics_comparison}) by reflecting the decline in image quality, but also produces more differentiated scores at major severity increments such as levels 5, 10, 15, and 20. 
This discriminative power and the inverse correlation between DreamSim's scores and corruption severity are valuable features of an NVS evaluation metric. 

In the cases where corruptions include a rotational component \textemdash \emph{i.e}\onedot, rotation and warp \textemdash scores increase as severity increases past a certain level due to the image approaching its original orientation. 
This reflects that the metrics identify when the corrupted image is maximally misaligned with the reference image. 
Additionally, a slight increase in score is noted in early contrast corruption levels due to initial application of high contrast that is subsequently lowered as additional contrast changes are applied.

The results of global corruption benchmarking on the ImageNet-C dataset are shown in Figure \ref{fig:ImageNetC_Corruption_Results}. 
This demonstrates that over a significantly increased sample size of 750k variants across a wider array of image classes, DreamSim's scores consistently decrease as image quality declines. 
DreamSim maintains a gradual score reduction in 10 of 12 corruption types. 
The exceptions to this behavior are reflected in irregular score trends for the remaining two corruption types: fog and elastic transform. 
Fog introduces a score plateau, which aligns with fog's observed tendency to have little perceptual change across severity levels, depicted in Figure 10. 
Elastic transform introduces rotation, affecting scores similarly to rotation and warp, as discussed above. 

Comparatively, PSNR shows little score discriminability between severities across every corruption type except brightness. 
SSIM and LPIPS perform more consistently with human perception, but output less distinctive scores than DreamSim between the extrema of the corruption levels. 
\subsection{Local Corruptions}
\begin{figure}
  \centering
  \includegraphics[width=0.90\linewidth]{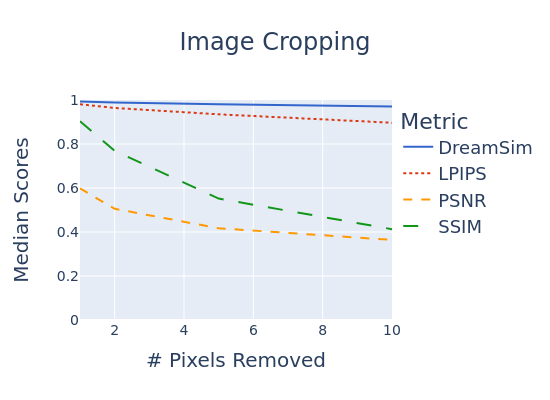}
  \caption{Median normalized scores of each metric as pixels are cropped along the outside edges of images.}
  \label{fig:PixelCrop_results}
\end{figure}

\subsection{Foreground/Background Corruption}
\begin{figure*}[t!]
  \begin{subfigure}{\textwidth}
    \centering
    \includegraphics[width=0.9\textwidth]{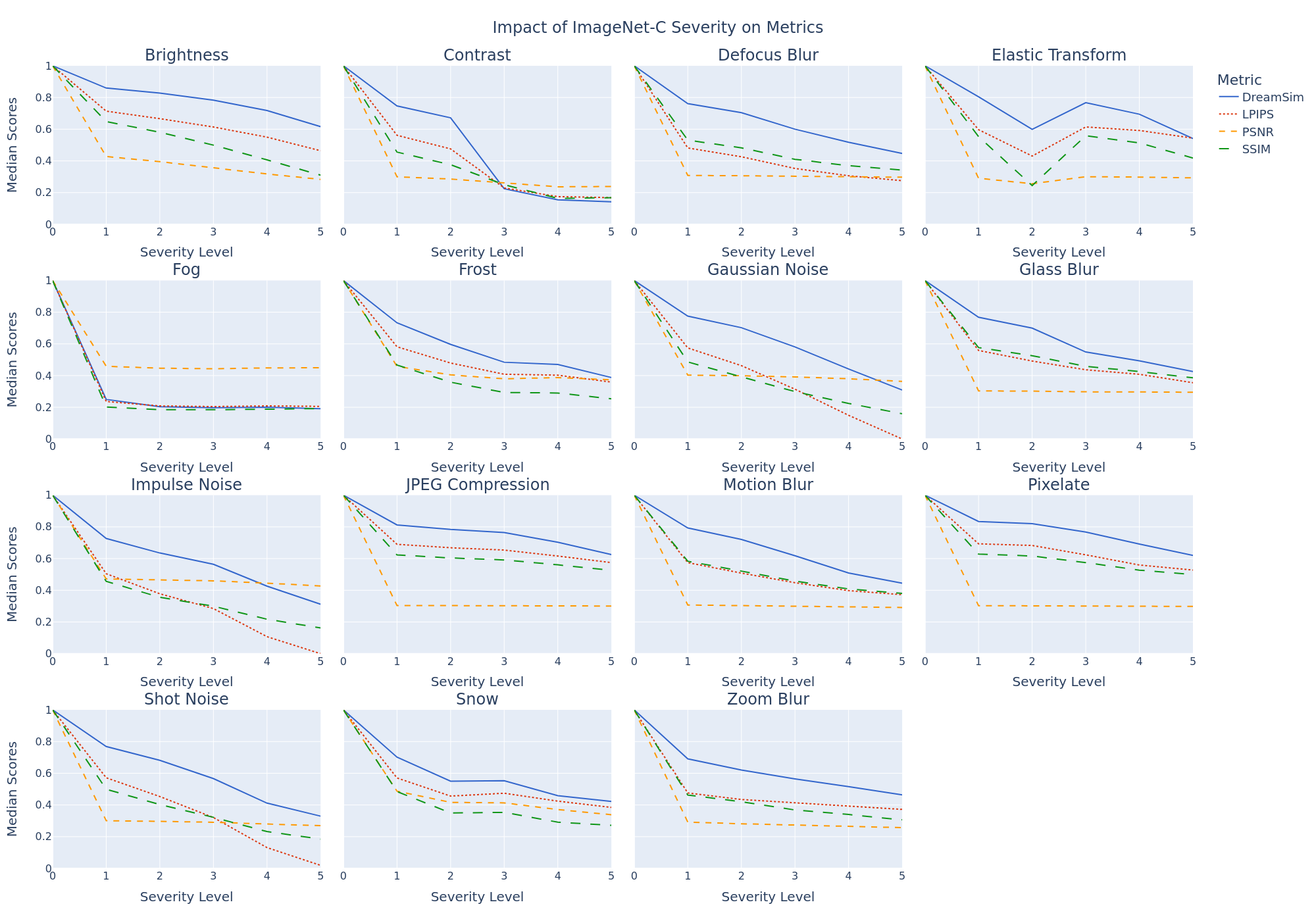}
    \caption{Figure \thefigure\ (a) Median scores across each severity level in the ImageNet-C dataset. All metric scores are normalized to the same scale 0 to 1, where 1 indicates an identical image pair.}
    \label{fig:ImageNetC_Corruption_Results}
  \end{subfigure}
  \bigskip
  \begin{subfigure}{\textwidth}
    \centering
    \includegraphics[width=0.9\textwidth]{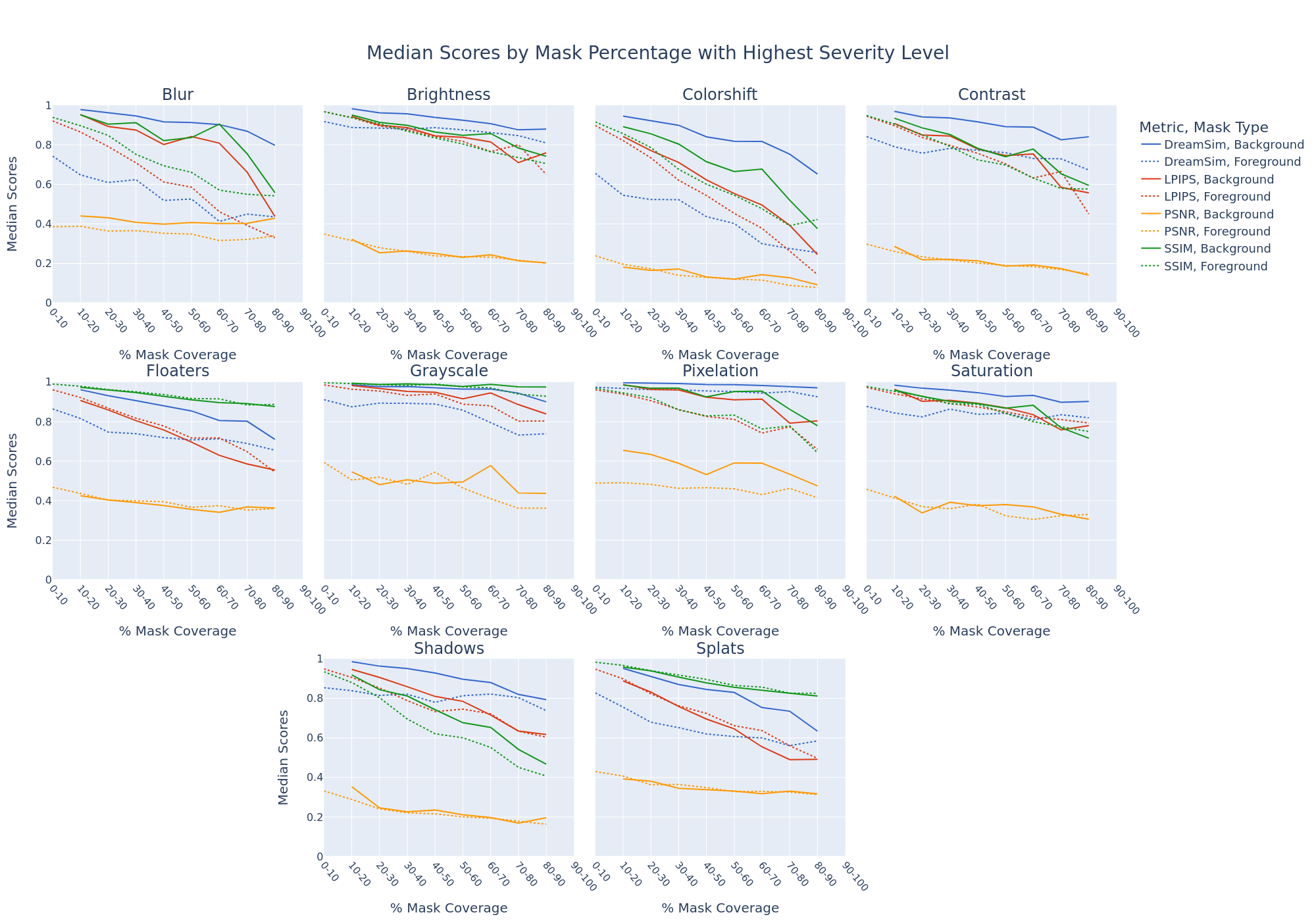}
    \caption{Figure \thefigure\ (b) Median scores resulting from foreground and background level 5 corruptions on the ImageNet data subset, for each corruption type, as corruption mask coverage increases. DreamSim and SSIM are normalized to the same 0 to 1 scale.}
    \label{fig:ForeBackground_Results}
  \end{subfigure}
\end{figure*}

The results in Figure \ref{fig:ForeBackground_Results} confirm that not only does the DreamSim score inversely correlate with increasing corruption size, but also that corruptions to the foreground impact the score more severely than corruptions to the background. 
For PSNR, SSIM, and LPIPS, there is little discrimination between foreground and background corruptions and scores are primarily influenced by the size of the corruption in the image. 
Although increased sensitivity to foreground object clarity and decreased sensitivity to background corruptions reflects an algorithmic bias, this trait is present in human visual systems \cite{hvs_bias}, indicating that DreamSim's evaluation criteria align more closely with human judgement.
\subsection{Imperceptible Image Cropping}
When observing NVS rendered images holistically, small, imperceptible artifacts are not meaningful to the viewer and would ideally have a minimal effect on a human-centric, applications-focused evaluation metric. 
The results of testing this quality of each metric via small, imperceptible cropping of images can be seen in Figure \ref{fig:PixelCrop_results}. 
In this experiment, DreamSim scores are the least impacted by minimal corruptions that do not affect overall scene understanding. 
In contrast, pixel-level metrics such as SSIM and PSNR degrade significantly after only a few pixels are removed, indicating that their scores are dramatically impacted by corruptions that have no impact to the holistic understanding of the image.
\section{Conclusions}

Currently, evaluating novel view generated imagery in research settings relies heavily on low-level image similarity metrics, such as SSIM, PSNR, and LPIPS. 
However, our experiments illustrate that these classical metrics are less meaningful for holistic evaluation of scene content within NVS renders, as evidenced by benchmarking these metrics on corrupted images. 
Specifically, in real-world applications with sparse input data, DreamSim is robust in scoring NVS renders in a predictable and differentiable manner. 

Lastly, our tests indicate DreamSim is less sensitive to corruptions imperceptible by humans, such as pixel-level changes found in NVS renders. 
DreamSim or a future NVS-specifically trained version of DreamSim should be used for general evaluation of NVS models to quantify render alignment with human judgment.

\FloatBarrier
\cleardoublepage
\twocolumn

\section*{Acknowledgements}
This work was accomplished in support of the Intelligence Advanced Research Projects Activity (IARPA), Office of Analysis.
This document was first produced for the U.S. Government under contract 2020-20090800401-017. 
The views and conclusions contained herein are those of the authors and should not be interpreted as necessarily representing the official policies or endorsements, expressed or implied, of IARPA or the U.S. Government or any of the authors host affiliations.

\texttt{@}2024 The MITRE Corporation. 
ALL RIGHTS RESERVED.

{\small
\bibliographystyle{ieee_fullname}
\bibliography{egbib}
}

\end{document}